\documentclass[prl,aps,twocolumn,superscriptaddress]{revtex4-2}
\usepackage{amsmath}
\usepackage{amssymb}
\usepackage{graphicx}
\usepackage{hyperref}
\usepackage{diagbox}
\usepackage{makecell}
\usepackage{booktabs}
\usepackage{mathtools}
\usepackage{tikz}
\usepackage[utf8]{inputenc}
\usepackage{xcolor}
\usepackage{CJKutf8}

\newcommand{\removed}[1]{\sout{Removed.}}

\usepackage{amsfonts}
\usepackage{amssymb}
\usepackage{listings} 
\usepackage[most]{tcolorbox} 

\newtcolorbox{promptbox}[1][]{
  colback=blue!5!white, 
  colframe=blue!75!black, 
  arc=2mm, 
  boxrule=0.5pt, 
  title=#1, 
  fonttitle=\bfseries,
  breakable, 
  enhanced, 
  top=2mm, bottom=2mm, left=2mm, right=2mm 
}

\newtcolorbox{verbatimbox}[1][]{
  colback=gray!10!white, 
  colframe=gray!75!black, 
  arc=2mm, 
  boxrule=0.5pt, 
  title=#1, 
  fonttitle=\bfseries,
  breakable, 
  enhanced, 
  top=2mm, bottom=2mm, left=2mm, right=2mm, 
  fontupper=\ttfamily 
}

\begin{document}
\begin{CJK*}{UTF8}{song}

  \title{Learning-at-Criticality in Large Language Models for Quantum Field Theory and Beyond}
  \author{Xiansheng Cai (\CJKfamily{gbsn}蔡贤盛)}
  \altaffiliation{These authors contributed equally to this paper.}
  \affiliation{Institute of Theoretical Physics, Chinese Academy of Sciences, Beijing 100190, China}
  \author{Sihan Hu (\CJKfamily{gbsn}胡思寒)}
  \altaffiliation{These authors contributed equally to this paper.}
  \affiliation{Hefei National Laboratory, University of Science and Technology of China, Hefei 230088, China}
  \author{Tao Wang (\CJKfamily{gbsn}王韬)}
  \altaffiliation{Corresponding authors. Email: taowang@iphy.ac.cn; huangyuan@dp.tech; panzhang@itp.ac.cn; yjdeng@ustc.edu.cn; chenkun@itp.ac.cn}
  \affiliation{Department of Physics, University of Massachusetts, Amherst, MA 01003, USA}
  \affiliation{Institute of Physics, Chinese Academy of Sciences, Beijing 100190, China}
  \author{Yuan Huang (\CJKfamily{gbsn}黄远)}
  \altaffiliation{Corresponding authors. Email: taowang@iphy.ac.cn; huangyuan@dp.tech; panzhang@itp.ac.cn; yjdeng@ustc.edu.cn; chenkun@itp.ac.cn}
  \affiliation{DP Technology, Beijing 100080, China}
  \author{Pan Zhang (\CJKfamily{gbsn}张潘)}
  \altaffiliation{Corresponding authors. Email: taowang@iphy.ac.cn; huangyuan@dp.tech; panzhang@itp.ac.cn; yjdeng@ustc.edu.cn; chenkun@itp.ac.cn}
  \affiliation{Institute of Theoretical Physics, Chinese Academy of Sciences, Beijing 100190, China}
  \affiliation{School of Fundamental Physics and Mathematical Sciences, Hangzhou Institute for Advanced Study, UCAS, Hangzhou 310024, China}
  \affiliation{Hefei National Laboratory, Hefei 230088, China}
  \author{Youjin Deng (\CJKfamily{gbsn}邓友金)}
  \altaffiliation{Corresponding authors. Email: taowang@iphy.ac.cn; huangyuan@dp.tech; panzhang@itp.ac.cn; yjdeng@ustc.edu.cn; chenkun@itp.ac.cn}
  \affiliation{Hefei National Laboratory, University of Science and Technology of China, Hefei 230088, China}
  \affiliation{Hefei National Laboratory for Physical Sciences at the Microscale and Department of Modern Physics, University of Science and Technology of China, Hefei 230026, China}
  \author{Kun Chen (\CJKfamily{gbsn}陈锟)}
  \altaffiliation{Corresponding authors. Email: taowang@iphy.ac.cn; huangyuan@dp.tech; panzhang@itp.ac.cn; yjdeng@ustc.edu.cn; chenkun@itp.ac.cn}
  \affiliation{Institute of Theoretical Physics, Chinese Academy of Sciences, Beijing 100190, China}
  \date{\today}
  \begin{abstract}
    Fundamental physics often confronts complex symbolic problems with few guiding exemplars or established principles. While artificial intelligence (AI) offers promise, its typical need for vast datasets to learn from hinders its use in these information-scarce frontiers. We introduce learning at criticality (LaC), a reinforcement learning (RL) scheme that tunes Large Language Models (LLMs) to a sharp learning transition, addressing this information scarcity. At this transition, LLMs achieve peak generalization from minimal data, exemplified by 7-digit base-7 addition—a test of nontrivial arithmetic reasoning. To elucidate this peak, we analyze a minimal concept-network model (CoNet) designed to capture the essence of how LLMs might link tokens. Trained on a single exemplar, this model also undergoes a sharp learning transition. This transition exhibits hallmarks of a second-order phase transition, notably power-law distributed solution path lengths. At this critical point, the system maximizes a ``critical thinking pattern" crucial for generalization, enabled by the underlying scale-free exploration. This suggests LLMs reach peak performance by operating at criticality, where such explorative dynamics enable the extraction of underlying operational rules. We demonstrate LaC in quantum field theory: an 8B-parameter LLM, tuned to its critical point by LaC using a few exemplars of symbolic Matsubara sums, solves unseen, higher-order problems, significantly outperforming far larger models. LaC thus leverages critical phenomena, a physical principle, to empower AI for complex, data-sparse challenges in fundamental physics.
  \end{abstract}
  \maketitle
\end{CJK*}

\emph{Introduction}---Artificial intelligence (AI) has accelerated scientific discovery, yet its primary successes are in data-rich domains leveraging pattern recognition, a capability closely aligned with intuitive ``System 1" thinking~\cite{stanovich_individual_2000, kahneman_thinking_2011, hua-zhang-2022-system,wei_chain--thought_2023,kabraTwoBrainsOne2023}.
A distinct class of frontier scientific problems, particularly in theoretical physics, presents a different challenge: they often necessitate deriving complex analytical solutions through extended abstract, ``System 2" reasoning, yet the very nature of these frontiers means training data for AI models is inherently limited~\cite{zhangSelfConsistencyTrainingDensityFunctionalTheory2024,hormannMachineLearningDataDriven2025,mcgiffOvercomingDataScarcity2025,pangDeepPathOvercomingData2025}, seemingly placing them beyond the reach of conventional AI reliant on vast statistical correlations.

This chasm between current AI strengths and the needs of theoretical physics is starkly evident. In Quantum Electrodynamics (QED), the electron's anomalous magnetic moment ($a_e$) acts as a critical test of the Standard Model. While numerical evaluations of $a_e$ coefficients are known to high precision (e.g., up to the fifth loop), their complete analytical derivation---essential for deep theoretical insight---is achieved only to the third loop after decades of effort~\cite{schwingerQuantumElectrodynamicsMagneticMoment1948,laportaAnalyticalValueElectron1996,aoyamaTenthorderElectronAnomalous2015,aoyamaTenthOrderQEDContribution2012,fanMeasurementElectronMagnetic2023}. Similarly, in many-electron systems in condensed matter, understanding phenomena like high-temperature superconductivity relies on analytically mastering the Fermi-surface complexities in Feynman diagrams. Here too, numerical methods like Diagrammatic Monte Carlo (DiagMC) provide crucial estimates for higher-order terms~\cite{diagmc1,diagmc2,diagmc2010,diagmckozik}, but the analytical solution of even low-order diagrams remains a challenging task. For fields demanding generalizable symbolic reasoning from a few solved instances, AI models traditionally associated with ``System 1" appear ill-equipped.

However, recent advancements offer a new perspective. Large Language Models (LLMs), particularly when augmented with Reinforcement Learning (RL), are beginning to exhibit capabilities that transcend simple pattern matching~\cite{vaswani_attention_2023,shao_deepseekmath_2024,deepseek-ai_deepseek-r1_2025,drgrpo,yu_dapo_2025,qwen2.5,qwen3,teamKimiK15Scaling2025}. RL enables LLMs to actively explore problem spaces, learn from feedback on their generated reasoning pathways, and refine strategies—processes that foster more coherent, goal-oriented, and multi-step thought, intriguingly reminiscent of ``System 2" cognition. This opens a promising, albeit challenging, avenue for AI to assist in domains demanding true analytical depth.

While LLMs augmented by RL show promise for ``System 2"-like reasoning, deploying them effectively in frontier science requires navigating inherent challenges.
Firstly, the highly specialized nature of these problems means they are often statistical outliers to an LLM's general pre-training~\cite{hormannMachineLearningDataDriven2025,mcgiffOvercomingDataScarcity2025,pangDeepPathOvercomingData2025,zhangSelfConsistencyTrainingDensityFunctionalTheory2024}.
This misalignment can lead to unreliable outputs or ``hallucinations"~\cite{huangSurveyHallucinationLarge2025,liuEnhancingMathematicalReasoning2025,yuBenchmarkingReasoningRobustness2025} when precise symbolic manipulation is critical, diminishing the utility of off-the-shelf models.
A natural corrective is targeted RL fine-tuning, tailoring the LLM to the specific nuances of the cutting-edge problem.
Yet, this essential fine-tuning step itself encounters a profound obstacle: the very frontier nature that necessitates such specialization also implies an extreme scarcity of existing data suitable for this RL process.
This creates a central dilemma: if the indispensable RL fine-tuning must operate with exceptionally few examples, can these advanced AI models genuinely acquire the algorithmic understanding and robust generalization capabilities characteristic of ``System 2" reasoning~\cite{wang_reinforcement_2025,gaoOneshotEntropyMinimization2025}? Addressing this question is crucial for determining AI's true potential in advancing fundamental science.

In this Letter, we resolve this central dilemma by introducing ``learning at criticality" (LaC), a learning scheme inspired by critical phenomena in physics. LaC precisely guides an LLM~\cite{qwen2.5} via RL to a narrow `critical' training phase. At this threshold, analogous to a phase transition, the LLM attains strong problem-solving capabilities and optimal generalization. This targeted training to criticality imparts to the LLM genuine, generalizable algorithmic understanding from minimal data---even a single exemplar---circumventing the data-dependency that has historically limited conventional AI in numerous scientific domains. Consequently, LaC enables AI to address complex, abstract theoretical problems in fundamental science previously considered intractable due to data scarcity.

\begin{figure}[h!]
  \centering
  \includegraphics[width=\columnwidth]{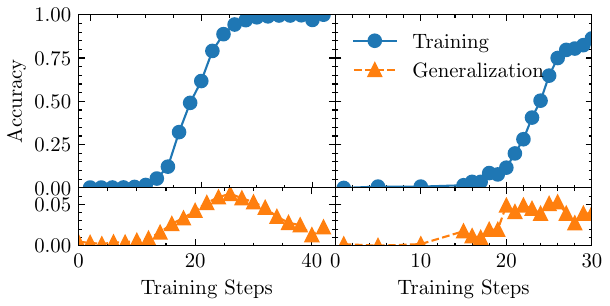}
  \caption{
    Critical learning from a single training example.
    (Left) Training a Qwen2.5-7B model on one 7-digit base-7 addition. Training accuracy (blue circles) shows a sharp transition. Generalization to unseen additions (orange triangles) peaks precisely at this critical point before overfitting.
    (Right) Similar phenomenon for a Qwen3-8B model trained on Matsubara frequency summation (2-loop sunrise self-energy diagram). Generalization to other unseen 2-loop diagrams is maximized at the critical learning point.
  }
  \label{fig:a7b7single}
\end{figure}

\begin{figure}[h!]
  \centering
  \includegraphics[width=0.9\columnwidth]{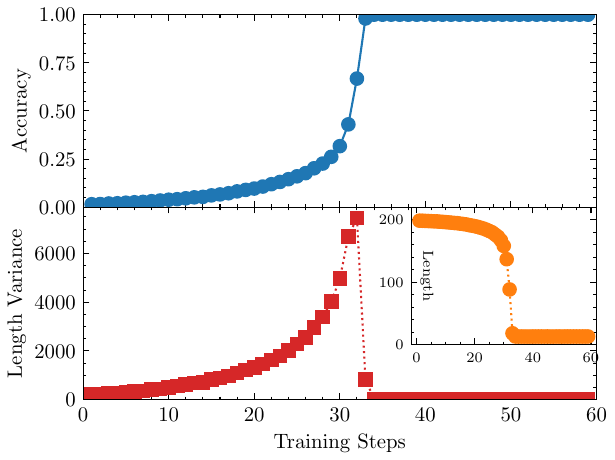}
  \caption{
    Training dynamics of the minimal concept-network model (CoNet). The figure shows the accuracy, average response length, and the response length's variance of the minimal model on the training problem, plotted against training steps. The accuracy (blue) increases and the average response length (orange) decreases in a sigmoidal manner. Concurrently, the response length's variance (red) exhibits a lambda-shape discontinuity at the learning transition, mirroring the behavior of specific heat at the lambda point marking the normal-to-superfluid helium phase transition.
  }
  \label{fig:toymodel_training}
\end{figure}

\begin{figure}[h!]
  \centering
  \includegraphics[width=\columnwidth]{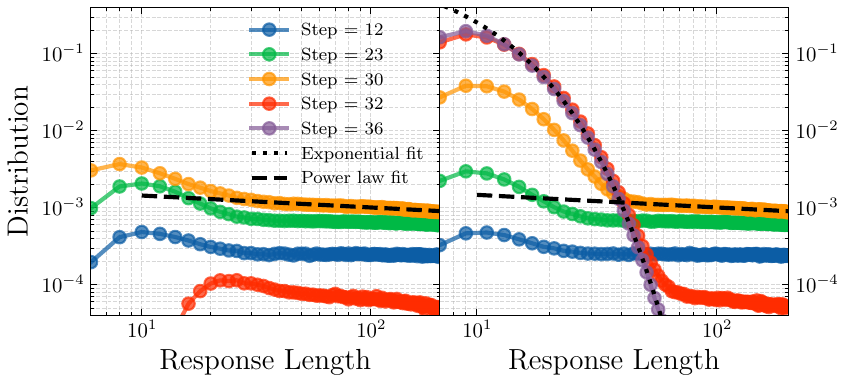}
  \caption{
    Distinct reasoning dynamics: critical power-law search transitions to post-convergence exponential exploration. Toy model reasoning response length distributions $P(L)$ across training epochs. During the critical learning transition (e.g., step 30, orange), long exploratory responses exhibit characteristic power-law decay $P(L) \sim L^{-\gamma}$ with $\gamma \approx 0.16$ (dashed fit, left panel; also evident for early-stage odd paths, right panel), a signature of scale-invariant critical search. Post-transition, as the policy converges (e.g., to a 7-step optimal odd path, step 34, dark grey), local perturbations around this path display exponential decay $P(L) \sim e^{-\alpha L}$ (dotted fit, right panel).
    These distinct scaling regimes characterize the evolution from broad, critical exploration to refined exploitation, crucial for learning. Distributions are truncated at the maximum allowed response length.
  }
  \label{fig:toymodel_step_loglog}
\end{figure}

We first establish LaC by training an LLM on a single instance of 7-digit, base-7 addition, observing a peak in its generalization capability at a precise training stage (Fig. \ref{fig:a7b7single}, left panel). To understand the physics underlying this peak, we propose the concept-network model (CoNet). This minimal model abstracts LLM reasoning, where cohesive sequences of autoregressively generated tokens form ``concepts". The LLM's problem-solving is viewed as a stochastic traversal within an implicit network of these concepts, with RL training (e.g., a variation of Group Relative Policy Optimization (GRPO)~\cite{yu_dapo_2025, drgrpo, shao_deepseekmath_2024,deepseek-ai_deepseek-r1_2025}) adjusting inter-concept transition probabilities. Our CoNet embodies this as a Markovian walk on a random graph of such ``concepts", where transitions are learned via reinforcement to find paths from question to answer. Remarkably, this simplified CoNet reproduces a sharp learning transition exhibiting hallmarks of a second-order phase transition: problem-solving accuracy (Fig.\ref{fig:toymodel_training}, upper panel) increases sigmoidally, while reasoning path length variance diverges, signaling critical fluctuations. Crucially, near this critical point, path lengths become power-law distributed ($L^{-\gamma}$, Fig.~\ref{fig:toymodel_step_loglog}), indicating an emergent ``critical thinking pattern" of diverse strategic exploration.

Finally, we apply LaC to the symbolic Matsubara frequency summation in Feynman diagrams---a challenging problem in finite-temperature many-body quantum field theory (QFT)~\cite{abrikosov1963methods,mahan2000many,nieto_evaluating_1995,espinosa_evaluation_2009}. Remarkably, fine-tuning an 8-billion parameter LLM~\cite{qwen3} to its critical point using only low-order diagrams enables it to learn the symbolic procedure and solve unseen, more complex diagrams (Fig.~\ref{fig:a7b7single}, right panel), outperforming models with nearly two orders of magnitude more parameters (Tab. \ref{tab:accuracyofmatsum}). Our work establishes LaC as a data-efficient strategy for AI-driven discovery in theoretical physics and suggests that emergent reasoning in AI can be understood as a critical phenomenon.

\emph{Learning at Criticality: A Phenomenon}--- Before presenting a theoretical model, we empirically demonstrate that effective learning from sparse data, while achievable, critically depends on the training regime. We tasked a Qwen2.5-7B model \cite{qwen2.5}, initially unable to perform 7-digit base-7 addition, with learning this procedure from a \emph{single} problem instance. This specific task was chosen to probe algorithmic reasoning. Its multi-step, rule-based nature (requiring sequential digit-by-digit processing and explicit base-7 carrying rules) serves as a strong proxy for deliberative ``System 2" cognition. Furthermore, its presumed rarity in pre-training corpora, particularly when contrasted with arithmetic in common bases like 2 or 10, significantly minimizes the likelihood of the model recalling a memorized solution rather than learning the underlying algorithm.

We trained the model on this one example using Direct-Advantage Policy Optimization (DAPO) \cite{yu_dapo_2025}, which is a variant of the GRPO algorithm\cite{shao_deepseekmath_2024,deepseek-ai_deepseek-r1_2025}. It operates by having the LLM generate multiple potential reasoning paths for a given problem. These paths are evaluated, and their scores, relative to the group average, guide policy updates to the LLM's parameters. This process strengthens transitions on paths yielding above-average rewards and weakens those on paths performing below average, thereby preferentially guiding the policy towards effective reasoning strategies.

As shown in Fig.~\ref{fig:a7b7single} (left panel), the training accuracy on the single problem instance (blue circles) displays a sharp, sigmoidal transition, indicating a sudden acquisition of the solution. This transition is not perfectly singular, likely due to multiple valid reasoning paths (e.g., different ways to handle carrying operations), but the qualitative feature is clear. More importantly, when we tested the model's performance on 128 new, unseen 7-digit base-7 addition problems, its generalization accuracy (orange triangles) peaked precisely in the vicinity of this learning transition. Although this initial peak generalization rate was approximately $7\%$---a modest absolute value, yet significant given it stems from a single training example—we further validated that by carefully continuing the training within this critical regime, the model progressively enhanced its capability to solve unseen instances~\cite{supmat} correctly. As training continued past this point (overfitting), the model's general ability declined, even as its performance on the single training example remained perfect. This finding is the cornerstone of our ``learning at criticality" proposal: it demonstrates the remarkable capacity of an LLM to acquire genuinely generalizable, algorithmic understanding from even a single exemplar, provided the training navigates and sustains it within a transient, optimal learning state.

\emph{CoNet Model for Learning Transitions}---LLMs generate text autoregressively, predicting subsequent tokens based on prior context. When an LLM predicts a sequence of tokens where each next token has nearly $100\%$ certainty, we consider these tokens to form a cohesive unit, which we abstract as a ``concept". LLM reasoning can then be viewed as a stochastic traversal---akin to a random walk---within an underlying network of these concepts. GRPO training step effectively acts as an external ``tuning parameter" that modifies the transition probabilities within the LLM's implicit concept network, optimizing pathways from the ``question concepts" to the ``answer concepts".

To model this process and understand the physics of LaC, we propose the CoNet. In this minimal model, the LLM's abstract concept space is represented as a K-regular random graph with N nodes (concepts). A reasoning task is modeled as finding a path from a source node Q (question) to a target node A (answer), constrained to a maximum path length of $L_\text{max}=200$. The LLM's token generation is simplified to a Markovian walk on this graph, where the transition probability from concept $i$ to a neighboring concept $j$ is
\begin{equation}
  \pi_\theta (j|i) = \frac{\theta_{ij}}{\sum_{k \in \text{neighbors}(i)} \theta_{ik}},
\end{equation}
where $\theta_{ij} \in [0,1)$ are learnable parameters representing transition strengths. For a given Q-A pair, $M=10^4$ reasoning paths (indexed by $m$) are sampled. Each path receives a reward, and its advantage $A_m$ (relative to the average reward) guides the update of $\theta_{ij}$ via a GRPO-variant rule~\cite{shao_deepseekmath_2024, drgrpo, yu_dapo_2025} $\Delta\theta_{ij} \propto \sum_m A_m \nabla_{\theta_{ij}} \log\pi_{\theta}(j|i)$
This reinforces transitions on above-average paths.
  {It is important to clarify that CoNet is a minimalist abstraction of the emergent reasoning process, not the LLM's architecture. The model's concepts (nodes) correspond to stable, low-entropy token sequences, while its learnable transitions (links) represent the probabilistic choices made at high-entropy decision points during generation.}

Simulations of CoNet ($N=8000$ nodes, $K=5$) reveal a sharp learning transition (Fig.~\ref{fig:toymodel_training}), {a behavior that proves robust across various model parameters as shown in the Supplemental Material~\cite{supmat}}.
The accuracy (fraction of successful paths, serving as an order parameter) displays a steep, sigmoidal increase. Concurrently, the variance of the reasoning path lengths exhibits a pronounced peak at the transition, analogous to diverging susceptibility at a critical point in physical systems. These features are hallmarks of a continuous phase transition and mirror the empirical learning peak.

The microscopic origin of this transition is revealed in the hybrid nature of the reasoning path distribution $P(L)$ at criticality. As shown in Fig.~\ref{fig:toymodel_step_loglog} (around step 30), the distribution is effectively a superposition of two distinct search strategies. A peak at short lengths signifies local exploration around an emergent optimal path, a behavior that consolidates into a purely exponential decay ($P(L) \sim e^{-\alpha L}$) post-transition (step 36). Coexisting with this is a pronounced power-law tail, $P(L) \sim L^{-\gamma}$ ($\gamma \approx 0.16$), a canonical signature of scale-free, critical phenomena {(see Supplemental Material~\cite{supmat} for similar results with a larger maximum response length)}. The prominence of this exploratory, power-law mode is maximized at the transition, directly causing the large path-length variance (susceptibility). This ``critical thinking pattern" represents a diverse, long-range exploratory search, and its coexistence with efficient exploitation is the hallmark of the learning transition, enabling the discovery of generalizable strategies before the system collapses into a single, non-exploratory state.

\emph{AI for QFT}--- To demonstrate LaC's potential for addressing data-scarce~\cite{hormannMachineLearningDataDriven2025,mcgiffOvercomingDataScarcity2025,pangDeepPathOvercomingData2025,zhangSelfConsistencyTrainingDensityFunctionalTheory2024}, symbolically complex problems in theoretical physics, we apply it to a challenging representative task: the symbolic evaluation of Matsubara frequency sums~\cite{nieto_evaluating_1995,espinosa_evaluation_2009}. This multi-step, algorithmically rich procedure is foundational to finite-temperature QFT~\cite{abrikosov1963methods,mahan2000many} and methods like DiagMC for many-electron problems~\cite{diagmc1,diagmc2,diagmc2010,diagmckozik,diagmclatest,detdiagmc,vdiagmc1,vdiagmc2,burkeTorchAmiGeneralizedCPU2025,leblancDynamicResponseElectron2022a,taheridehkordiAlgorithmicMatsubaraIntegration2019}. An example is the 2-loop sunrise self-energy diagram (Eq.\ref{eq:sunrise}), whose symbolic solution involves contour integration and complex analysis.
\begin{align}
  \label{eq:sunrise}
  \vcenter{\hbox{\includegraphics{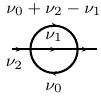}}} = & \sum_{\nu_0,\nu_1}\dfrac{1}{\nu_0+\varepsilon_0}\dfrac{1}{\nu_1+\varepsilon_1}\dfrac{1}{\nu_0+\nu_2-\nu_1+\varepsilon_2} \notag                                                   \\
  =                                                               & \oint_{C}\frac{dz^{\prime}}{2\pi i}\frac{f(z^{\prime})}{z^{\prime}+\varepsilon_1}\oint_{C}\frac{dz}{2\pi i}\frac{f(z)}{z+\varepsilon_0}\frac{1}{z+\nu_2-z^{\prime}+\varepsilon_2}
\end{align}
where $\nu$ are fermionic Matsubara frequencies, $\epsilon$ are energy parameters, and $f(z)$ is the Fermi-Dirac distribution. This task is an ideal LaC testbed due to its structured complexity, availability of low-order diagrams for sparse training, and verifiable analytical solutions, allowing us to probe for emergent algorithmic understanding.

The well-defined, symbolic nature of this procedure makes it an ideal testbed for LaC. Specifically: (i) low-order diagrams offer sparse training instances; (ii) complexity scales with diagram order, naturally testing generalization; and (iii) exact analytical solutions permit unambiguous verification. Our objective is to investigate if an LLM, trained via staged LaC on minimal examples, can acquire the algorithmic reasoning inherent in this task, highlighting LaC's potential for data-efficient learning of complex physics procedures.

Our training employed a staged LaC strategy with a Qwen3-8B model \cite{qwen3} using GRPO. Phase 1: LaC training on only tree-level and 1-loop Matsubara sums. This not only enabled solutions for these simpler diagrams but also induced initial generalization to higher-order diagrams decomposable into these learned components. Notably, while the base model had zero success on the challenging sunrise self-energy diagram (Eq.~\ref{eq:sunrise})---a 2-loop graph with nested frequencies---this first LaC phase elevated its performance to a non-zero baseline. Phase 2: This crucial improvement enabled a second LaC phase focused specifically on the sunrise diagram. As shown in Fig.~\ref{fig:a7b7single} (right panel), the model again exhibited a sharp learning transition for this complex task. Critically, at this new critical point, the ability gained from mastering the sunrise diagram is generalized to other 2-loop topologies (e.g., polarization, vertex functions), where performance was previously negligible, confirming the LaC hypothesis. Phase 3: Finally, starting from this enhanced critical state, we incorporated all 2-loop diagrams into the LaC training until a critical point was reached for this comprehensive set.

\begin{table}[h]
  \centering
  \begin{tabular}{c|cccc}
    \toprule
    \diagbox{Model}{Accuracy(\%)}{Problem}
                               & \makecell{1-loop}
                               & \makecell{2-loop}
                               & \makecell{3-loop}
                               & \makecell{4-loop}                                                     \\
    \midrule
    DeepSeek-R1-0120(671B)     & 12.5              & 10.0            & 1.1            & 0.4            \\
    Qwen3-32B                  & 47.8              & 25.5            & 1.6            & 0.6            \\
    Qwen3-8B                   & $45.0$            & $18.4$          & $0.9$          & $0.2$          \\
    \hline
    \makecell{Qwen3-8B-1-loop} & $90.2$            & $25.5$          & 3.3            & 0.8            \\
    \makecell{Qwen3-8B-2-loop} & $\mathbf{97.5}$   & $\mathbf{56.9}$ & $\mathbf{9.5}$ & $\mathbf{1.7}$ \\
    \bottomrule
  \end{tabular}
  \caption{Critical learning transition enables generalization for Matsubara sums. Qwen3-8B, fine-tuned on problems up to 2-loop complexity, exhibits a sharp increase in success rate. This critical learning phase facilitates robust generalization to unseen 3-loop overlapped sums.
    Consequently, this fine-tuned 8B model markedly outperforms significantly larger base models (Qwen3-32B, DeepSeek-R1 671B) on this task, indicating data-efficient acquisition of algorithmic reasoning.}
  \label{tab:accuracyofmatsum}
\end{table}

The results are shown in Table~\ref{tab:accuracyofmatsum}. The base Qwen3-8B model's success is modest for 1-loops ($45.0\%$) and 2-loops ($18.4\%$), and minimal for 3-loops ($0.9\%$) and 4-loops ($0.2\%$). The latter non-zero values reflect occasional solutions to simple, decomposable higher-order diagrams, not general complex reasoning. In contrast, the fully fine-tuned model (Qwen3-8B-2-loop, post-Phase 3) greatly improved accuracy on its 2-loop training set. Crucially, it then achieved strong accuracy on complex 3-loop problems, despite no explicit training on them, and generalized to 4-loop cases. This LaC-trained model far surpasses its base and larger models (DeepSeek-R1, Qwen3-32B), underscoring LaC's efficiency. This success demonstrates that navigating critical points enabled the LLM to develop an emergent, algorithmic grasp of Matsubara summation.

\emph{Conclusion}---We addressed the challenge of ``System 2'' AI reasoning in data-scarce science by introducing ``learning at criticality'' (LaC), where LLM fine-tuning via RL induces a critical phase transition towards generalizable algorithmic understanding from minimal data.

This LaC principle was first shown with 7-digit base-7 addition: generalization from a single example peaked at the sharp learning transition. Our minimal physical model \emph{CoNet} indicates that it is a continuous phase transition, revealing power-law scaling in reasoning paths---a signature of emergent ``critical thinking". Critically, applying LaC to symbolic Matsubara frequency summation, an 8-billion parameter LLM, trained on a few low-order diagrams, learn the procedure and solves unseen, more complex diagrams, substantially outperforming models nearly two orders of magnitude larger.
  {Important open questions remain regarding how the LaC phenomenon scales with LLM size. While our CoNet model provides a tractable framework for exploring network parameters, the equivalent scaling experiments on large language models are computationally expensive. We therefore leave this investigation for future work.}

Critically, this LaC-induced phase transition occurs in an implicit concept network linked to ``System 2" reasoning, differentiating it from grokking phenomena which occur during pre-training or supervised fine-tuning and are associated with ``System 1" pattern matching~\cite{powerGrokkingGeneralizationOverfitting2022,linAnalysisFrameworkUnderstanding2025}. LaC thus provides a novel mechanism for cultivating advanced AI capabilities.

This work provides both a physical framework for understanding emergent AI reasoning and a practical, data-efficient strategy for applying it to data-scarce domains like theoretical physics.
The implications of LaC extend deeply into domains like QFT, where navigating immense symbolic complexity with limited examples of optimal strategies is paramount~\cite{songExplainableAIassistedOptimization2025, hippelRefiningIntegrationbyPartsReduction2025}. Beyond theoretical physics, LaC provides a pathway towards AI systems as powerful scientific collaborators, equipped to tackle complex symbolic manipulations and accelerate discovery across diverse frontiers of knowledge where data is sparse but the demand for deep reasoning is high.

\emph{Acknowledgements}--- The authors thank Haijun Zhou, Lei Wang, Linfeng Zhang, and Pengcheng Hou for inspiring discussions.
K. C. and X. C. are supported by the National Key Research and Development Program of China under Grant No.
2024YFA1408604 and the National Natural Science
Foundation of China under Grants No. 12047503 and
No. 12447103. P.Z. is supported by Project 12325501 of the National Natural Science Foundation of China. Y.D. and S.H. are supported by the National Natural Science Foundation of China under Grants No. 12275263, the Innovation Program for Quantum Science and Technology under Grant No. 2021ZD0301900, and the Natural Science Foundation of Fujian Province of China under Grant No. 2023J02032.

\bibliographystyle{myref}
\bibliography{format_ref}

\appendix
\onecolumngrid 
\section*{Abstracting LLM Reasoning into a Concept Network}
\label{sec:appendix_conet}
\renewcommand{\theequation}{S\arabic{equation}}
\renewcommand{\thefigure}{S\arabic{figure}}
\renewcommand{\thetable}{S\arabic{table}}
\setcounter{equation}{0}
\setcounter{figure}{0}
\setcounter{table}{0}

\begin{figure}[h!]
  \centering
  \includegraphics[width=0.85\columnwidth]{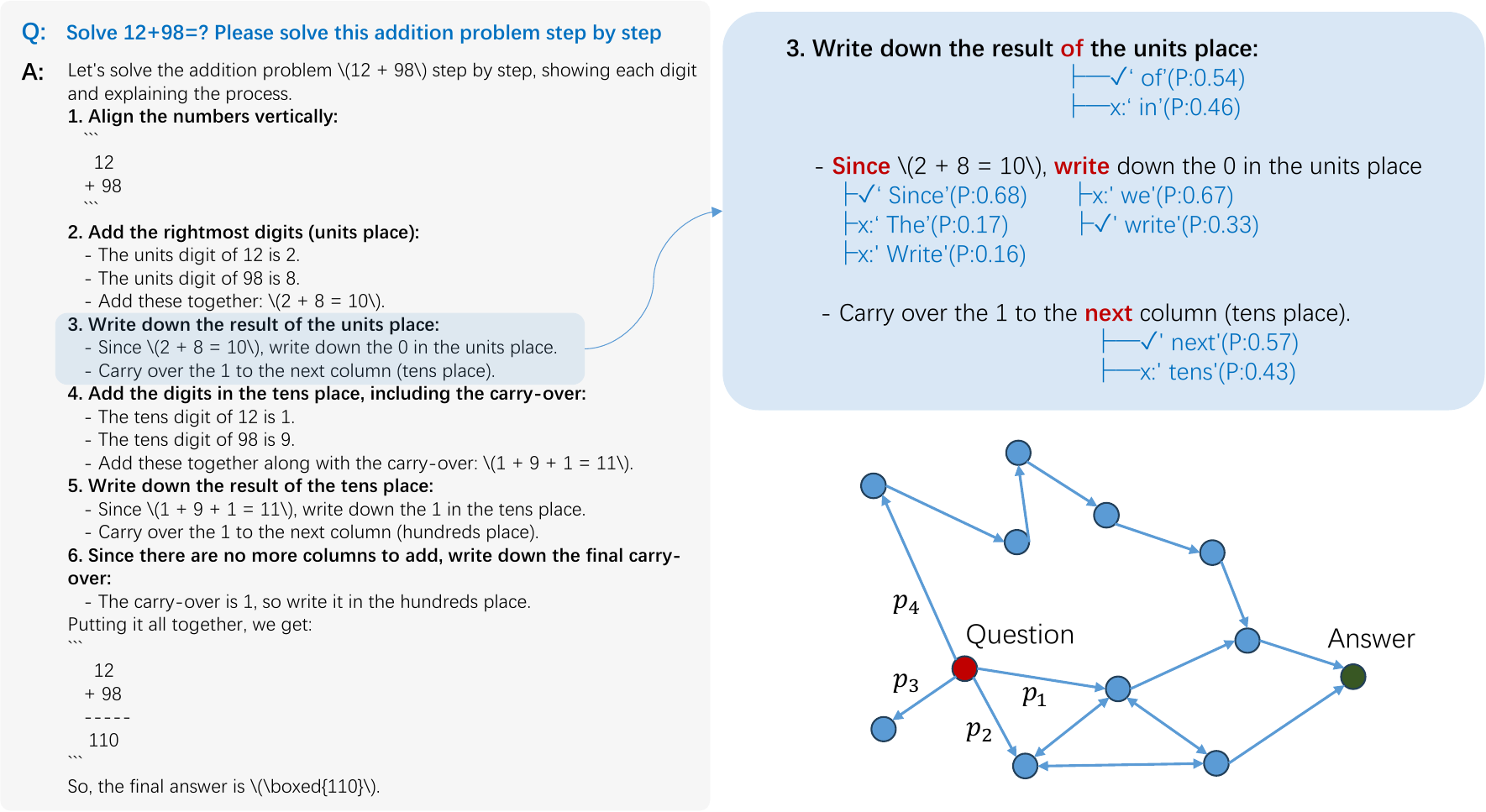}
  \caption{\textbf{Abstraction from LLM token generation to a concept network.} The figure illustrates the abstraction process using an example from the Qwen2.5-7B-Instruct model. (Left) The model's step-by-step reasoning. (Top Right) A detailed view of a \textbf{high-entropy decision point}, where the probability distribution over the next token is broad. The high-probability candidates (e.g., P(`Since`)=0.68 vs. P(`The`)=0.17) are the competing ``forking tokens'' that define the branches of the network. (Bottom Right) A schematic of the resulting CoNet, where reasoning is a stochastic path from ``Question'' to ``Answer'' node(s).}

  \label{fig:conet}
\end{figure}

In the main text, we model the LLM's reasoning process as a stochastic traversal on an abstract concept network (CoNet). This appendix details how this abstraction is derived from the LLM's underlying autoregressive token generation, as illustrated in Fig.~\ref{fig:conet}. The LLM does not generate text with uniform uncertainty; instead, its process is characterized by sequences of high-confidence, low-entropy tokens punctuated by key \textbf{decision points} of high uncertainty. We define a ``concept'' as a coherent sequence of tokens generated between two such decision points. Each of these stable text chunks, such as ``3. Write down the result'', forms a single node in our abstract network.

The transitions between these concept nodes are formed at the decision points. At these junctures, the LLM must make a meaningful choice that determines the subsequent reasoning path. This aligns with the concept of high-entropy ``forking tokens'' recently identified as critical for reinforcement learning in LLMs \cite{wang2025beyond}. Crucially, both our analysis and their concurrent work find that these tokens, which steer the reasoning process, often serve as abstract logical connectors (e.g., ``since'', ``next'', ``thus'') and may lack specific domain content themselves. As shown in Fig.~\ref{fig:conet}, the probabilities assigned to these competing forking tokens represent the directed, weighted edges of our network. These branching probabilities (e.g., P(``next'')=0.57 vs. P(``tens'')=0.43) are the transition probabilities in our model.

Consequently, an LLM's complete solution to a problem corresponds to a single sampled path through this vast, implicit concept network. The schematic in the figure illustrates this view, where multiple potential reasoning paths ($p_1, p_2, \dots$) emanate from the initial ``Question'' node(s), each representing a different chain of thought. Some paths successfully navigate the network to reach the correct ``Answer'' nodes, while many others terminate in incorrect states or fail to terminate within a given token quota.

The GRPO training process implicitly manipulates this network structure. By rewarding successful paths, it reinforces the constituent transitions, increasing the probabilities of edges that form correct reasoning steps. This dynamically reshapes the transition landscape, guiding the model to learn reliable problem-solving algorithms. Therefore, the CoNet model presented in the main text serves as a minimal, idealized representation of this abstracted network, allowing us to isolate and study the fundamental physics of the learning transition without the confounding complexities of the full LLM architecture.



\section*{Reinforcement Learning Training Details}
\label{sec:appendix_rl_details}
\renewcommand{\theequation}{S\arabic{equation}}
\renewcommand{\thefigure}{S\arabic{figure}}
\renewcommand{\thetable}{S\arabic{table}}
\setcounter{equation}{0}
\setcounter{figure}{0}
\setcounter{table}{0}

In this section, we provide further details on the Reinforcement Learning (RL) fine-tuning procedures employed to train our Large Language Models (LLMs) following the ``learning at criticality'' (LaC) paradigm introduced in the main Letter. All RL experiments were executed using the \texttt{verl} framework~\cite{verl_ref,verl_github}, an open-source library designed for LLM reinforcement learning which facilitates algorithms such as Group Relative Policy Optimization (GRPO)~\cite{shao_deepseekmath_2024} and its variants, including Direct-Advantage Policy Optimization (DAPO)~\cite{yu_dapo_2025}.

\subsection{CoNet Model and Training Protocol}
\begin{figure}
  \centering
  \includegraphics[width=0.4\linewidth]{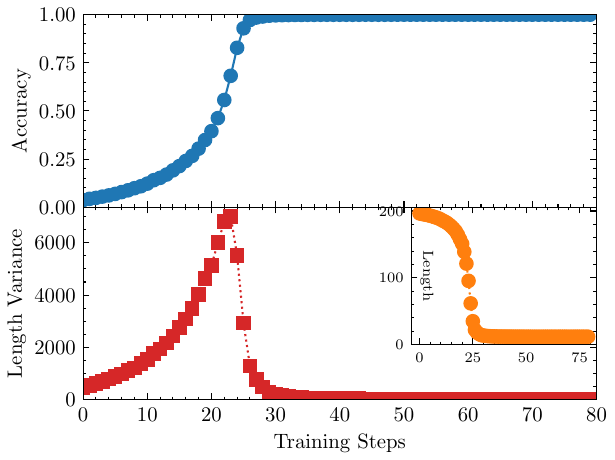}
  \includegraphics[width=0.4\linewidth]{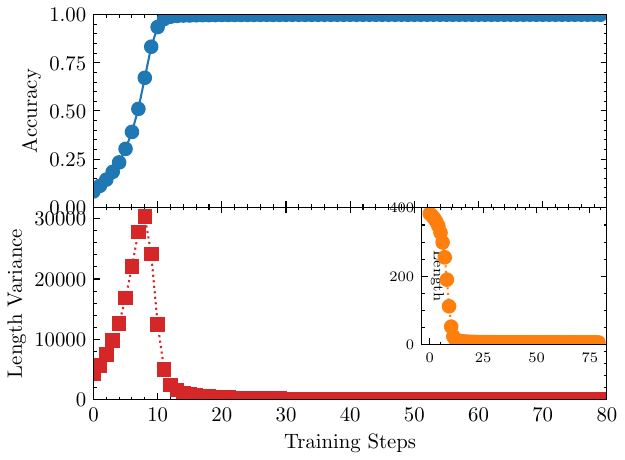}

  \includegraphics[width=0.4\linewidth]{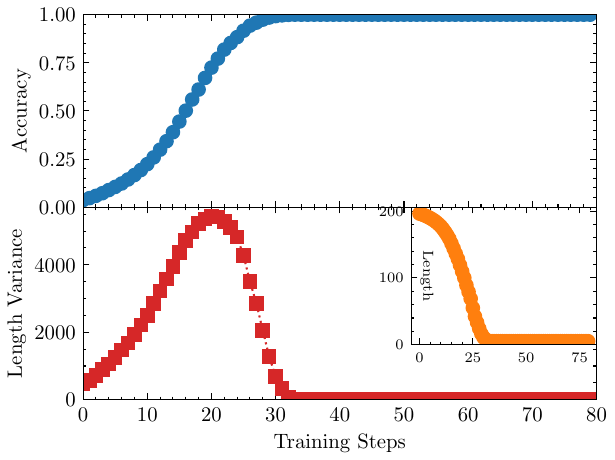}
  \includegraphics[width=0.4\linewidth]{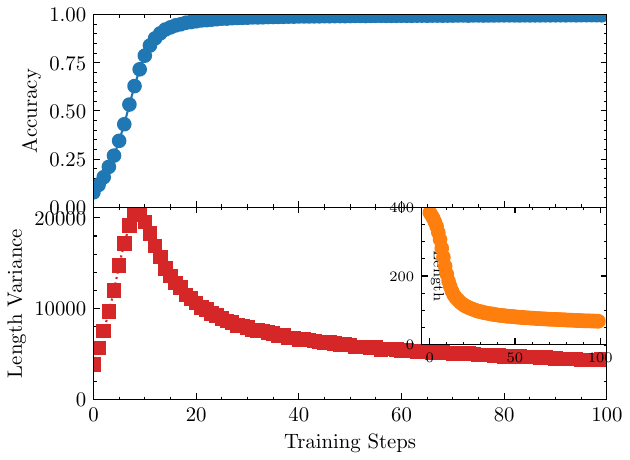}

  \includegraphics[width=0.4\linewidth]{figures/training_logs_test5_inset.pdf}
  \includegraphics[width=0.4\linewidth]{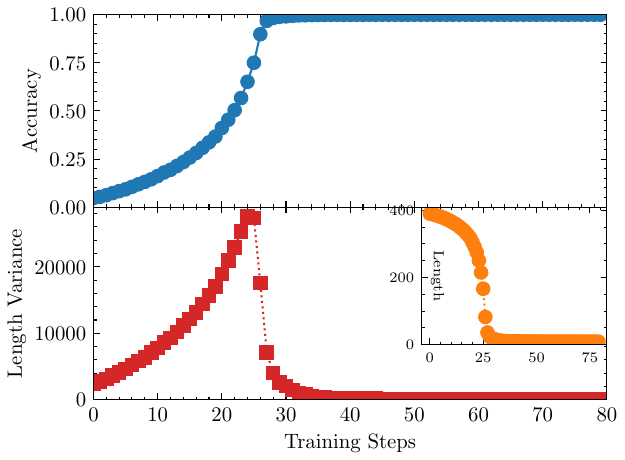}

  \includegraphics[width=0.4\linewidth]{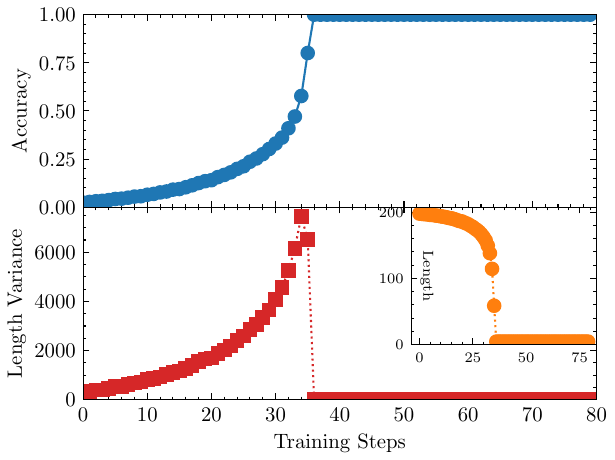}
  \includegraphics[width=0.4\linewidth]{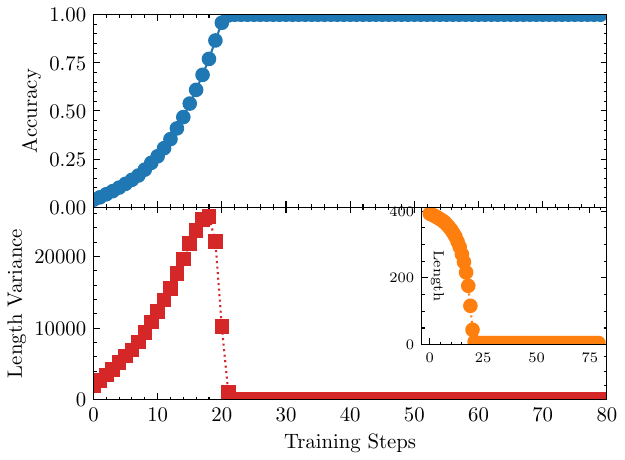}
  \caption{
    Additional training dynamics of the CoNet model for various parameters, consistent with the behavior shown in FIG.~\ref{fig:toymodel_training}. From top to bottom, the rows correspond to networks with node numbers and degrees of $(N,K)=(4000,5),(4000,10),(8000,5),$ and $(8000,10)$. The left and right panels depict simulations with maximum response lengths $L_\text{max}$ of 200 and 400, respectively. All results show a similar learning transition, but systems with a smaller node number, larger degree and maximum response length exhibit more significant finite-size effects, resulting in a more smeared transition.
  }
  \label{fig:conet_training_logs}
\end{figure}

\begin{figure}
  \centering
  \includegraphics[height=4.6cm]{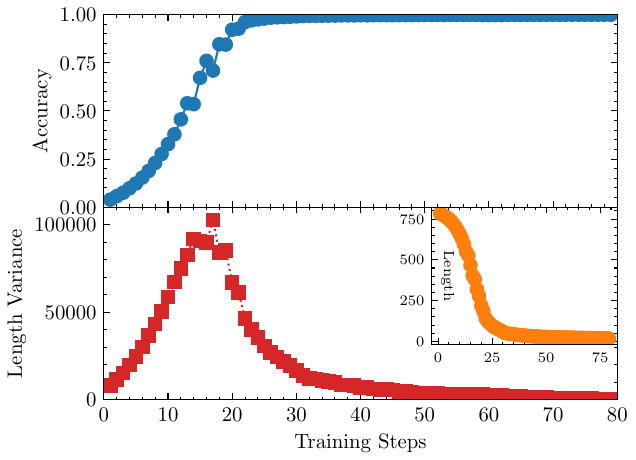}
  \includegraphics[height=4.6cm]{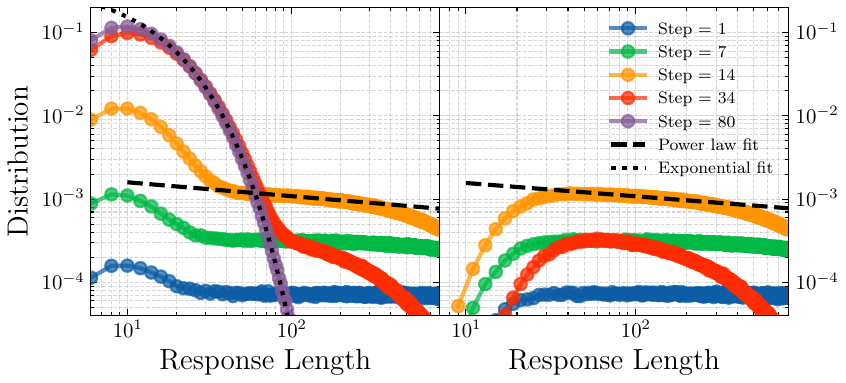}
  \caption{
    CoNet training dynamics with an extended maximum response length. This figure reproduces the analysis of FIG.~\ref{fig:toymodel_training} and FIG.~\ref{fig:toymodel_step_loglog} but with the maximum response length increased to $L_\text{max}=800$, compared to $L_\text{max}=200$ in the main text. (Left Panel) The model exhibits the same characteristic learning transition, though it appears less sharp due to more pronounced finite-size effects. (Right Panel) The response length distribution at the critical point (step 15, orange curve) again shows a power-law tail, $P(L)\sim L^{-\gamma}$ with $\gamma\approx0.16$, in consistent with FIG.~\ref{fig:toymodel_step_loglog}. The finite-size effect is also evident here, as the distribution for longer response lengths begins to show exponential decay before the shorter length part, indicating a more gradual convergence process.
  }
  \label{fig:toymodel_extend}
\end{figure}

This section provides the specific training protocol for the supplementary CoNet simulations shown in FIG.~\ref{fig:conet_training_logs} and~\ref{fig:toymodel_extend}, which were performed to verify the robustness of the LaC phenomenon.
For these simulations, the policy gradient is estimated from 64,000 sampled reasoning paths per training step. The policy parameters are updated using a learning rate of 0.3; to maintain numerical stability, they are subsequently clipped to the range [-0.2, 0.3]. The results presented in FIG.~\ref{fig:conet_training_logs} demonstrate that the sharp learning transition is a robust feature, persisting across the various network sizes, degrees, and maximum response lengths tested. FIG.~\ref{fig:toymodel_extend} further investigates the system with a significantly extended maximum response length ($L_\text{max}=800$), corroborating the critical dynamics observed in the main text while illustrating the expected increase in finite-size effects on the transition's sharpness.

\subsection{7-Digit Base-7 Addition}

The task of 7-digit base-7 addition was selected to assess the LLM's capacity for acquiring algorithmic reasoning from extremely sparse data, specifically a single problem exemplar.

The Qwen2.5-7B model~\cite{qwen2.5} served as the base LLM for this task. Fine-tuning was conducted using Direct-Advantage Policy Optimization (DAPO)~\cite{yu_dapo_2025}, recognized for its computational efficiency as a GRPO variant.

The model's training was based on a singular instance of a 7-digit base-7 addition problem. The input presented to the LLM consisted of a \texttt{system\_prompt} and a \texttt{user\_prompt}, structured as follows:

\begin{promptbox}[System Prompt]
  \texttt{You are Qwen, created by Alibaba Cloud. You are a helpful assistant. The assistant first thinks about the reasoning process in the mind and then provides the user with the answer. The reasoning process is enclosed within <think> </think> tag, i.e., <think> reasoning process here </think> answer here. Now the user asks you to solve a math problem. Try to think step by step. After thinking, when you finally reach a conclusion, clearly state the final result within \textbackslash{}boxed\{\} tag, for example, \textbackslash{}boxed\{1\}.}
\end{promptbox}

\begin{promptbox}[User Prompt (Base-7 Addition)]
  \texttt{Add the two base-7 numbers 6454502 and 1214210. First, pad both numbers with leading zeros so they have the same number of digits. Then, start from the rightmost (least significant) digit and perform the addition digit by digit from right to left. For each digit position, add the corresponding digits of 6454502 and 1214210 (treating missing digits in the shorter number as 0). Add any carry-over from the previous step. Record the sum digit (sum modulo 7) and carry-over (sum divided by 7) for the next position. Continue this process until you have processed all digits of the longer number AND resolved any remaining carry-over (e.g., if adding 66 + 1 in base-7, you must account for the final carry to get 100 in base-7). Present the final result as a base-7 number, including all digits. Perform the calculation digit by digit in the 7-base system without converting to base-10. Put your final answer in \textbackslash{}boxed\{...\}.}
\end{promptbox}

The ground truth solution for this specific training instance (6454502 + 1214210 in base-7) is \texttt{\textbackslash{}boxed\{11002012\}}.

A binary reward signal (+1 or 0) was employed. A reward of +1 was granted if the LLM's final answer, enclosed within the \texttt{\textbackslash{}boxed\{\}} tags, precisely matched the known ground truth solution. Otherwise, a reward of 0 was assigned. The explicit reasoning process within the \texttt{<think> </think>} tags was not directly factored into the reward calculation but was encouraged by the prompt structure; its quality was observed to improve as training progressed.

The primary objective was to train the model until it consistently solved the single training exemplar and subsequently to evaluate its generalization capability on a set of 128 new, unseen 7-digit base-7 addition problems. As depicted in Fig.~\ref{fig:a7b7single} (left panel) of the main Letter, fine-tuning on this solitary example induced a sharp, sigmoidal increase in training accuracy. Significantly, the model's generalization performance on unseen problems peaked precisely at this critical learning transition, thereby empirically validating the LaC hypothesis for this algorithmic reasoning task.
  {To further enhance performance, we continued training from this peak-generalization checkpoint using an iterative approach. The process involved repeatedly switching to a new, difficult problem and training the model only until its generalization rate on the unseen dataset peaked again. This "nudging" procedure was repeated for several cycles, which successfully kept the model in the critical state, avoided overfitting, and raised the overall generalization rate to approximately $30\%$.}

To illustrate the model's state before and after LaC training, we present representative outputs. The base Qwen2.5-7B model typically failed to perform base-7 addition correctly, often making procedural errors. For instance, when prompted to add \texttt{3304461} and \texttt{1555230} in base-7, a base model might produce an incorrect sequence of operations and an erroneous result like \texttt{\textbackslash{}boxed\{1355266\}} (the correct sum is \texttt{5163021}):
\begin{verbatimbox}[Model Output Before Training (Base-7 Addition)]
  Let's add 3304461 and 1555230 in base-7. Both numbers have 7 digits, so no padding is necessary. Starting from the rightmost digit, we have: 6 + 0 = 6 5 + 3 = 8 (carry-over 1, sum 1) 4 + 2 = 6 (carry-over 1, sum 2) 4 + 5 = 12 (carry-over 1, sum 5) 3 + 5 = 12 (carry-over 1, sum 5) 0 + 1 = 1 (carry-over 0, sum 1) 3 + 0 = 3 (carry-over 0, sum 3) 1 + 0 = 1 (carry-over 0, sum 1) The sum in base-7 is 1355266. \boxed{1355266}
\end{verbatimbox}

In contrast, after LaC fine-tuning on the primary training exemplar (\texttt{6454502 + 1214210}), the model demonstrated correct algorithmic execution not only on the training exemplar, but also on the distinct problem \texttt{3304461 + 1555230}, producing the following type of output:
\begin{verbatimbox}[LaC-Trained Model Output on Training Exemplar (Base-7 Addition)]
  <think>
  To add the two base-7 numbers 6454502 and 1214210, we first ensure they have the same number of digits by padding them with leading zeros if necessary. However, in this case, they already have the same number of digits (7 digits each).

  We will add them digit by digit from right to left, remembering to carry over any value greater than 6 (since it's base-7).

  Starting from the rightmost digit:\\
  - 2 + 0 = 2 (no carry)\\
  - 0 + 1 = 1 (no carry)\\
  - 5 + 2 = 7. In base-7, 7 is represented as 10, so we write down 0 and carry over 1.\\
  - 4 + 4 + 1 (carry) = 9. In base-7, 9 is represented as 12, so we write down 2 and carry over 1.\\
  - 5 + 1 + 1 (carry) = 7. In base-7, 7 is represented as 10, so we write down 0 and carry over 1.\\
  - 4 + 2 + 1 (carry) = 7. In base-7, 7 is represented as 10, so we write down 0 and carry over 1.\\
  - 6 + 1 + 1 (carry) = 8. In base-7, 8 is represented as 11, so we write down 1 and carry over 1.

  Since we have a carry-over after the leftmost digit, we add another digit to the left, making it 1.

  So the result is 11002012 in base-7.\\
  </think>

  \boxed{11002012}
\end{verbatimbox}
\begin{verbatimbox}[LaC-Trained Model Output on Generalization Problem (Base-7 Addition)]
  <think>
  To add the base-7 numbers 3304461 and 1555230, we first align them by padding with leading zeros so they have the same number of digits. This gives us: ``` 3304461 + 1555230 ```

  Now, we add from right to left, carrying over as necessary: \\
  1. 1 + 0 = 1 (no carry) \\
  2. 6 + 3 = 9. In base-7, 9 is 12 (since 9 = 7*1 + 2), so we write down 2 and carry over 1. \\
  3. 4 + 2 + 1 (carry) = 7. In base-7, 7 is 10 (since 7 = 7*1 + 0), so we write down 0 and carry over 1. \\
  4. 4 + 5 + 1 (carry) = 10. In base-7, 10 is 13 (since 10 = 7*1 + 3), so we write down 3 and carry over 1. \\
  5. 0 + 5 + 1 (carry) = 6. We write down 6. \\
  6. 3 + 5 = 8. In base-7, 8 is 11 (since 8 = 7*1 + 1), so we write down 1 and carry over 1. \\
  7. 3 + 1 + 1 (carry) = 5. We write down 5.

  Putting it all together, we get the result in base-7 as 5163021.
  </think>
  \\
  The final answer is \boxed{5163021}.
\end{verbatimbox}
This successful generalization to a new problem instance, involving correct application of the base-7 addition rules, underscores the acquisition of the underlying algorithm.

\subsection{Symbolic Matsubara Frequency Summation}

The symbolic evaluation of Matsubara frequency sums represents a significantly more complex task, drawn from quantum field theory, that is characterized by a multi-step, algorithmically rich symbolic manipulation procedure.

For this task, the Qwen3-8B model~\cite{qwen3} was utilized. The fine-tuning process was conducted using the DAPO algorithm~\cite{shao_deepseekmath_2024, deepseek-ai_deepseek-r1_2025, yu_dapo_2025}.

The training regimen employed a staged LaC strategy, utilizing a minimal set of low-order Matsubara sum problems as exemplars. The prompt provided to the LLM contained the Matsubara sum expression formatted in LaTeX, along with detailed instructions guiding its symbolic evaluation. An example of the user prompt for the 2-loop sunrise diagram (Eq.~\ref{eq:sunrise} in the main Letter) is as follows:
\begin{promptbox}[User Prompt (Matsubara Sum for the Sunrise Feynman Diagram)]
  Evaluate the Matsubara frequency summation for the following integrand from a Feynman diagram:
  $\sum_{\nu_0, \nu_1} \frac{1}{\nu_0+x_0}\frac{1}{\nu_0-\nu_1+\nu_2+x_2}\frac{1}{\nu_1+x_1}$,
  where $\nu_0$, $\nu_1$ are the independent internal fermionic Matsubara frequencies and $\nu_2$ is the independent external fermionic Matsubara frequency.\\
  \#\# Required Instructions\\
  1. Think step by step through your solution process\\
  2. Convert Matsubara Sum to Contour Integral:
  example:$$ \sum_{\nu_0} F(\nu_0) = \beta \oint \frac{dz}{2\pi i} F(z)f(z)$$,
  where the contour encloses counterclockwise all the poles of $F(z)$.
  The poles of $f(z)$ are at the imaginary axis and not enclosed by the contour.\\
  3. Identify all poles in the integrand\\
  4. Use contour integration and the residue theorem correctly\\
  5. Express your final answer as a complete LaTeX expression with the Fermi-Dirac distribution function: $f(x)\equiv\frac{1}{e^{\beta x}+1}$, where $\beta=\frac{1}{k_B T}$\\
  6. Place your final expression inside a \textbackslash{}boxed\{\} environment, preserving external frequencies and parameters.
\end{promptbox}

A binary reward system (+1 or 0) was implemented. The model received a reward of +1 if the final LaTeX expression rendered within the \texttt{\textbackslash{}boxed\{\}} environment was symbolically equivalent to the established correct analytical solution. Symbolic equivalence was rigorously verified using a computer algebra system (e.g., SymPy), involving parsing the LaTeX output and comparing canonical forms of the expressions.

As elaborated in the main Letter, the training protocol consisted of three distinct phases:
Phase 1 involved LaC training on tree-level and 1-loop Matsubara sum diagrams.
Phase 2 narrowed the LaC focus to the more complex 2-loop sunrise diagram (defined in Eq.~\ref{eq:sunrise} of the main Letter), initializing from the model state achieved at the end of Phase 1.
Phase 3 extended the LaC training to encompass all tree-level, 1-loop and 2-loop diagrams, commencing from the optimized model state from Phase 2.

Each phase was designed to guide the LLM towards a critical learning point specific to the complexity of the diagrams in that stage. Fig.~\ref{fig:a7b7single} (right panel) of the main Letter illustrates the characteristic sharp learning transition observed for the sunrise diagram during Phase 2. The developmental trajectory of the model's capabilities underscores the efficacy of this staged approach. Initially (pre-Phase 1), the base Qwen3-8B model's success rate on the challenging 2-loop sunrise diagram was undetectable. After the completion of Phase 1 (LaC training on 1-loop diagrams), the model achieved a modest but crucial success rate of approximately $0.01\%$ on the unseen sunrise diagram. This emergent, albeit minimal, capability was sufficient to enable effective RL fine-tuning in Phase 2, which focused specifically on the sunrise diagram. Upon reaching its critical learning point in Phase 2, the model's success rate for the sunrise diagram increased dramatically to approximately $50\%$. Concurrently, this mastery generalized to other unseen 2-loop topologies, such as polarization, 3-vertex, and 4-vertex diagrams, for which the success rate rose from negligible levels to around $5\%$. This new baseline, in turn, enabled the comprehensive LaC training of Phase 3 across all 2-loop diagrams. The final LaC-trained model, designated Qwen3-8B-2-loop (post-Phase 3), exhibited substantially improved accuracy on the 1-loop and 2-loop diagrams it was trained on. More importantly, it demonstrated robust generalization to unseen, higher-complexity 3-loop and 4-loop diagrams, as quantified in Table~\ref{tab:accuracyofmatsum} of the main Letter. This progression affirms the model's acquisition of the underlying symbolic summation algorithm through staged critical learning.

An example of the reasoning process generated by the fully LaC-trained Qwen3-8B-2-loop model for the Matsubara sum prompt shows a detailed, step-by-step derivation. This includes correct application of procedures such as contour integral conversion, pole identification, and residue calculations, culminating in a final symbolic expression:

\begin{verbatimbox}[Model Output After Training (Matsubara Sum for the Sunrise Feynman Diagram)]
  We are asked to evaluate the Matsubara frequency summation:

  $$
    \sum_{\nu_0, \nu_1} \frac{1}{\nu_0 + x_0} \frac{1}{\nu_0 - \nu_1 + \nu_2 + x_2} \frac{1}{\nu_1 + x_1}
  $$

  where $\nu_0, \nu_1$ are internal fermionic Matsubara frequencies, and $\nu_2$ is an external fermionic Matsubara frequency. We'll proceed step-by-step as per the instructions.

  ---

  \emph{Step 1: Understand the Structure of the Summation}

  This summation is over two independent Matsubara frequencies, $\nu_0$ and $\nu_1$, and the expression contains three terms in the denominator. The structure is:

  $$
    \sum_{\nu_0, \nu_1} \frac{1}{\nu_0 + x_0} \cdot \frac{1}{\nu_0 - \nu_1 + \nu_2 + x_2} \cdot \frac{1}{\nu_1 + x_1}
  $$

  We will first perform the sum over $\nu_0$, treating $\nu_1$ as a fixed parameter, and then over $\nu_1$.

  ---

  \emph{Step 2: Convert the Sum to a Contour Integral}

  Using the identity:

  $$
    \sum_{\nu} F(\nu) = \beta \oint \frac{dz}{2\pi i} F(z) f(z)
  $$

  where the contour encloses all poles of $F(z)$, and the Fermi-Dirac distribution is defined as:

  $$
    f(x) = \frac{1}{e^{\beta x} + 1}, \quad \text{with} \quad \beta = \frac{1}{k_B T}
  $$

  Let’s first perform the sum over $\nu_0$.

  ---

  \emph{Step 3: Fix $\nu_1$ and Sum over $\nu_0$}

  Let’s denote:

  $$
    S = \sum_{\nu_0} \frac{1}{\nu_0 + x_0} \cdot \frac{1}{\nu_0 - \nu_1 + \nu_2 + x_2}
  $$

  We now define:

  $$
    F(\nu_0) = \frac{1}{(\nu_0 + x_0)(\nu_0 - \nu_1 + \nu_2 + x_2)}
  $$

  So the sum becomes:

  $$
    \sum_{\nu_0} F(\nu_0) = \beta \oint \frac{dz}{2\pi i} F(z) f(z)
  $$

  Let’s simplify the denominator:

  $$
    F(z) = \frac{1}{(z + x_0)(z - \nu_1 + \nu_2 + x_2)} = \frac{1}{(z + x_0)(z + (\nu_2 - \nu_1 + x_2))}
  $$

  Let’s denote:

  $$
    a = x_0, \quad b = \nu_2 - \nu_1 + x_2
  $$

  So:

  $$
    F(z) = \frac{1}{(z + a)(z + b)}
  $$

  Now, we compute the contour integral:

  $$
    \sum_{\nu_0} F(\nu_0) = \beta \oint \frac{dz}{2\pi i} \frac{f(z)}{(z + a)(z + b)}
  $$

  This contour encloses all poles of $F(z)$, which are at $z = -a$ and $z = -b$, assuming these are not on the imaginary axis (which they are not for fermionic Matsubara frequencies). The poles are simple.

  So we can use the residue theorem:

  $$
    \sum_{\nu_0} F(\nu_0) = \beta \left[ \text{Res}_{z = -a} \left( \frac{f(z)}{(z + a)(z + b)} \right) + \text{Res}_{z = -b} \left( \frac{f(z)}{(z + a)(z + b)} \right) \right]
  $$

  Compute the residues:

  - At $z = -a$:

  $$
    \text{Res}_{z = -a} = \lim_{z \to -a} (z + a) \cdot \frac{f(z)}{(z + a)(z + b)} = \frac{f(-a)}{-a + b} = \frac{f(-a)}{b - a}
  $$

  - At $z = -b$:

  $$
    \text{Res}_{z = -b} = \lim_{z \to -b} (z + b) \cdot \frac{f(z)}{(z + a)(z + b)} = \frac{f(-b)}{-b + a} = \frac{f(-b)}{a - b}
  $$

  So:

  $$
    \sum_{\nu_0} F(\nu_0) = \beta \left( \frac{f(-a)}{b - a} + \frac{f(-b)}{a - b} \right) = \beta \left( \frac{f(-a)}{b - a} - \frac{f(-b)}{b - a} \right)
    = \beta \cdot \frac{f(-a) - f(-b)}{b - a}
  $$

  Recall:

  $$
    a = x_0, \quad b = \nu_2 - \nu_1 + x_2
  $$

  So:

  $$
    \sum_{\nu_0} F(\nu_0) = \beta \cdot \frac{f(-x_0) - f(-\nu_2 + \nu_1 - x_2)}{(\nu_2 - \nu_1 + x_2) - x_0}
  $$

  Now plug this back into the original sum:

  $$
    \sum_{\nu_0, \nu_1} \frac{1}{\nu_0 + x_0} \cdot \frac{1}{\nu_0 - \nu_1 + \nu_2 + x_2} \cdot \frac{1}{\nu_1 + x_1}
    = \sum_{\nu_1} \frac{1}{\nu_1 + x_1} \cdot \left[ \beta \cdot \frac{f(-x_0) - f(-\nu_2 + \nu_1 - x_2)}{(\nu_2 - \nu_1 + x_2) - x_0} \right]
  $$

  Let’s simplify the denominator:

  $$
    (\nu_2 - \nu_1 + x_2) - x_0 = \nu_2 + x_2 - x_0 - \nu_1
  $$

  So:

  $$
    \sum_{\nu_1} \frac{1}{\nu_1 + x_1} \cdot \left[ \beta \cdot \frac{f(-x_0) - f(-\nu_2 + \nu_1 - x_2)}{\nu_2 + x_2 - x_0 - \nu_1} \right]
    = \beta \cdot \left[ f(-x_0) - f(-\nu_2 + \nu_1 - x_2) \right] \cdot \sum_{\nu_1} \frac{1}{(\nu_1 + x_1)(\nu_2 + x_2 - x_0 - \nu_1)}
  $$

  Now we perform the sum over $\nu_1$:

  $$
    \sum_{\nu_1} \frac{1}{(\nu_1 + x_1)(\nu_2 + x_2 - x_0 - \nu_1)} = \sum_{\nu_1} \frac{1}{(\nu_1 + x_1)(\nu_2 + x_2 - x_0 - \nu_1)}
  $$

  Let’s denote:

  $$
    c = x_1, \quad d = \nu_2 + x_2 - x_0
  $$

  Then the expression becomes:

  $$
    \sum_{\nu_1} \frac{1}{(\nu_1 + c)(d - \nu_1)} = \sum_{\nu_1} \frac{1}{(\nu_1 + c)(d - \nu_1)} = \sum_{\nu_1} \frac{1}{(d - \nu_1)(\nu_1 + c)}
  $$

  Let’s define:

  $$
    F(\nu_1) = \frac{1}{(d - \nu_1)(\nu_1 + c)}
  $$

  Then again, use the same identity:

  $$
    \sum_{\nu_1} F(\nu_1) = \beta \oint \frac{dz}{2\pi i} \frac{f(z)}{(d - z)(z + c)}
  $$

  So:

  $$
    \sum_{\nu_1} \frac{1}{(d - \nu_1)(\nu_1 + c)} = \beta \left[ \text{Res}_{z = -c} + \text{Res}_{z = d} \right]
  $$

  Compute the residues:

  - At $z = -c$:

  $$
    \text{Res}_{z = -c} = \lim_{z \to -c} (z + c) \cdot \frac{f(z)}{(d - z)(z + c)} = \frac{f(-c)}{d + c}
  $$

  - At $z = d$:

  $$
    \text{Res}_{z = d} = \lim_{z \to d} (z - d) \cdot \frac{f(z)}{(d - z)(z + c)} = \frac{f(d)}{- (d + c)} = -\frac{f(d)}{d + c}
  $$

  So total:

  $$
    \sum_{\nu_1} \frac{1}{(d - \nu_1)(\nu_1 + c)} = \beta \left( \frac{f(-c)}{d + c} - \frac{f(d)}{d + c} \right) = \beta \cdot \frac{f(-c) - f(d)}{d + c}
  $$

  Recall:

  $$
    c = x_1, \quad d = \nu_2 + x_2 - x_0
  $$

  So:

  $$
    \sum_{\nu_1} \frac{1}{(d - \nu_1)(\nu_1 + c)} = \beta \cdot \frac{f(-x_1) - f(\nu_2 + x_2 - x_0)}{\nu_2 + x_2 - x_0 + x_1}
  $$

  Now putting it all together:

  $$
    \sum_{\nu_0, \nu_1} \frac{1}{\nu_0 + x_0} \cdot \frac{1}{\nu_0 - \nu_1 + \nu_2 + x_2} \cdot \frac{1}{\nu_1 + x_1}
    = \beta \cdot \left[ f(-x_0) - f(-\nu_2 + \nu_1 - x_2) \right] \cdot \beta \cdot \frac{f(-x_1) - f(\nu_2 + x_2 - x_0)}{\nu_2 + x_2 - x_0 + x_1}
  $$

  Wait — we have two $\beta$ factors from the two sums. Let’s retrace:

  The first sum over $\nu_0$ gave a factor of $\beta$, and the second sum over $\nu_1$ also gave a factor of $\beta$, so the total is:

  $$
    \sum_{\nu_0, \nu_1} \cdots = \beta^2 \cdot \left[ f(-x_0) - f(-\nu_2 + \nu_1 - x_2) \right] \cdot \frac{f(-x_1) - f(\nu_2 + x_2 - x_0)}{\nu_2 + x_2 - x_0 + x_1}
  $$

  But we can simplify the expression further by noticing that the term $- \nu_2 + \nu_1 - x_2$ can be rewritten as $\nu_1 - (\nu_2 + x_2)$, and the denominator is $\nu_2 + x_2 - x_0 + x_1 = \nu_2 + x_2 + x_1 - x_0$.

  So the final expression becomes:

  $$
    \boxed{ \beta^2 \cdot \frac{ \left[ f(-x_0) - f(\nu_1 - \nu_2 - x_2) \right] \left[ f(-x_1) - f(\nu_2 + x_2 - x_0) \right] }{ \nu_2 + x_2 + x_1 - x_0 } }
  $$

  ---

  \emph{Final Answer:}

  $$
    \boxed{ \beta^2 \cdot \frac{ \left[ f(-x_0) - f(\nu_1 - \nu_2 - x_2) \right] \left[ f(-x_1) - f(\nu_2 + x_2 - x_0) \right] }{ \nu_2 + x_2 + x_1 - x_0 } }
  $$
\end{verbatimbox}

\subsubsection{General Reinforcement Learning Hyperparameters}

For both tasks, unless specific tuning was performed to identify the LaC point, standard RL hyperparameters were utilized within the \texttt{verl} framework. These typically included a learning rate of $1 \times 10^{-6}$, a batch size of 8 prompts per update, and $M=32$ responses (rollouts) generated per prompt for the DAPO algorithm. This configuration results in 256 responses for the same single problem per each training step. The dynamic sampling was employed to ensure that the training always makes improvement. The AdamW optimizer was used. The critical aspect of the LaC methodology involves carefully monitoring the training dynamics to identify the onset of the sharp learning transition, which dictates the optimal point to halt or adjust training, rather than exhaustive hyperparameter optimization aimed solely at maximizing performance on a fixed dataset. Further details on the calibration of training duration for LaC are provided in the main Letter and its associated figures.

\end{document}